\documentclass[10pt,twocolumn]{article}

\usepackage[utf8]{inputenc} %unicode support
\usepackage{longtable}
\usepackage{graphicx}
\usepackage{epstopdf}
\usepackage[fleqn]{amsmath}
\usepackage{bm}
\usepackage{amsmath}
\usepackage{amssymb}
\usepackage{amsthm}
\usepackage{lmodern} %rai
\usepackage{fix-cm}  %rai
\usepackage{algorithm}
\usepackage{algorithmic}
\renewcommand{\algorithmicrequire}{\textbf{Input:}}
\renewcommand{\algorithmicensure}{\textbf{Output:}}

\newtheorem{proposition}{Proposition}[section]

\usepackage{color}
\definecolor{red}{rgb}{1,0,0}
\definecolor{blue}{rgb}{0,0,1}

\newcommand{\0}{{\bm{0}}}

\newcommand{\vh}{{\bm{h}}}

\newcommand{\vv}{{\bm{v}}}

\newcommand{\x}{{\bm{x}}}

\newcommand{\bE}{{\mathbb{E}}}

\newcommand{\vG}{{\bm{G}}}

\newcommand{\vH}{{\bm{H}}}
\newcommand{\vI}{{\bm{I}}}

\newcommand{\vM}{{\bm{M}}}
\newcommand{\bN}{{\mathbb{N}}}
\newcommand{\cN}{{\mathcal{N}}}

\newcommand{\vQ}{{\bm{Q}}}

\newcommand{\bR}{{\mathbb{R}}}

\newcommand{\vS}{{\bm{S}}}

\newcommand{\vV}{{\bm{V}}}

\newcommand{\vThet}{{\bm{\Theta}}}

\newenvironment{summary}{\begin{abstract}}{\end{abstract}}

\newcommand{\tslong}[1]{{#1}}
\newcommand{\tsshort}[1]{{}}

\newcommand{\defeq}{{:=}}
\newcommand{\coloneqq}{{:=}}
\newcommand{\eqqcolon}{{=:}}

\begin{document}

\twocolumn[{%
\begin{center}
  {\Large
    Mutual Kernel Matrix Completion}
\\
\vspace{1cm}
       {\large
         Tsuyoshi Kato${}^{\dagger,\ddagger,*}$,
         Rachelle Rivero${}^{\dagger}$}
\\
\vspace{1cm}
\begin{tabular}{lp{0.7\textwidth}}
${}^\dagger$ & 
Faculty of Science and Engineering, Gunma University, 
Kiryu-shi, Gunma, 326--0338, Japan.  
\\
${}^\ddagger$ &
Center for Informational Biology, Ochanomizu University, 
Bunkyo-ku, Tokyo,
112--8610, Japan. 
\end{tabular}
\end{center}
}]

%% \nocite{KatTsuAsa05a,RelNagKat16a}

\begin{summary}
With the huge influx of various data nowadays, extracting knowledge from them has become an interesting but tedious task among data scientists, particularly when the data come in heterogeneous form and have missing information. Many data completion techniques had been introduced, especially in the advent of kernel methods. However, among the many data completion techniques available in the literature, studies about mutually completing several incomplete kernel matrices have not been given much attention yet. In this paper, we present a new method, called Mutual Kernel Matrix Completion (MKMC) algorithm, that tackles this problem of mutually inferring the missing entries of multiple kernel matrices by combining the notions of data fusion and kernel matrix completion, applied on biological data sets to be used for classification task. We first introduced an objective function that will be minimized by exploiting the EM algorithm, which in turn results to an estimate of the missing entries of the kernel matrices involved. The completed kernel matrices are then combined to produce a model matrix that can be used to further improve the obtained estimates. An interesting result of our study is that the E-step and the M-step are given in closed form, which makes our algorithm efficient in terms of time and memory. After completion, the (completed) kernel matrices are then used to train an SVM classifier to test how well the relationships among the entries are preserved. Our empirical results show that the proposed algorithm bested the traditional completion techniques in preserving the relationships among the data points, and in accurately recovering the missing kernel matrix entries. By far, MKMC offers a promising solution to the problem of mutual estimation of a number of relevant incomplete kernel matrices.
\end{summary}

\section{Introduction}
\label{sec:Intro}
Analysis of biological data has been an interesting topic among researchers in the field, due to the challenges presented by biology being a data-rich science: the data are represented in various forms, and are painstakingly obtained. Most of the time, the data obtained are noisy and have missing information. The simplest technique in dealing with missing data is to delete the data points with missing entries; however, it reduces the data size and its statistical power \cite{Little}. For subsequent vector-based analysis, some missing data are imputed with zero (zero-imputation) or mean (mean-imputation) in advance. However, doing so dilutes the relationship among variables, which leads to underestimation of the variance \cite{Roth,Little}. On the other hand, some methods estimate with a regression model or EM algorithm.
The EM algorithm was introduced by Dempster \textit{et al.} as a general approach to compute the maximum likelihood estimates iteratively for incomplete data sets \cite{Dempster}. Each iteration consists of an expectation step (E-step) and a maximization step (M-step), hence the term \textit{EM algorithm}. 

Aside from missing information, another challenge faced by biological data is its heterogeneous representation. Biological data may come as strings (such as protein and genome sequences), microarrays (such as gene expression data), protein-protein interaction, or metabolic pathways. Simultaneous analysis of such data is difficult, unless they are converted into a single form. In Lanckriet \textit{et al.}~\cite{Lanckriet}, each data set is represented by a generalized similarity relationship between pairs of genes or proteins. This similarity relationship is defined by a kernel function; and data sets represented via a kernel function can be combined directly. The resulting matrix whose entries are the relationships defined by a kernel function is called a \textit{kernel matrix}. A huge literature about kernel methods and matrices is available: \cite{Tibshirani,Scholkopf,Vapnik,Bishop}. 

Lanckriet \textit{et al.}~\cite{Lanckriet} utilized the simultaneous analysis of kernel matrices in solving the problem of protein classification. They optimally combined multiple kernel representations by formulating the problem as a convex optimization problem that can be solved using semidefinite programming (SDP) techniques \cite{Lanckriet,LanckrietB,LanckrietC}. They had shown that a statistical learning algorithm performs better if trained from the integrated data than from a single data alone.

Integrated kernel matrices can also be used in the analysis of a data set with missing entries. For example, a supervised network inference among proteins can be formulated as a kernel matrix completion problem, whose missing entries can be inferred from multiple biological data types such as gene expression, phylogenetic profiles, and amino acid sequences \cite{Kato}. Moreover, when a data set has missing entries, its corresponding kernel matrix will be incomplete as well. Inferring the missing entries of the corresponding matrix rather than the data matrix itself has several advantages \cite{Kumar}: there are far fewer missing entries in the kernel matrix than in the actual data; and machine learning methods like support vector machines (SVMs) work with kernel matrices, to name a few. The completed kernel matrix, as well as the other relevant kernel matrices, can now be used in protein classification or protein network inference problems. Several kernel matrix completion methods that depend on other complete kernel matrices had been introduced: \cite{Tsuda,Kato,KKT}. However, it is not always the case that only one data set is incomplete. To date, simultaneous analysis and mutual completion of kernel matrices have not been extensively studied, but is now becoming an increasing interest among data scientists. Kumar \textit{et al.}~\cite{Kumar} considered the problem of deriving a single consistent similarity kernel matrix (with binary entries) from a set of self-consistent but incomplete similarity kernel matrices. They solved this problem by simultaneously completing the kernel matrices and finding their best linear combination via SVM. However, their algorithm involves an iterative optimization of each of the kernel matrices and weights via semidefinite program (SDP) solver, making their method very time- and memory-consuming. Another multiple kernel completion technique is introduced in \cite{Bhadra}. In their work, they only have one set of objects to be observed but some relationships between pairs of these objects are missing, generating a kernel matrix with missing rows and columns. They then generated multiple rearrangements of the said incomplete kernel matrix, thus generating multiple incomplete kernel matrices (multi-view setting) and posed the problem as multi-view completion problem. Their method then completes a kernel matrix with the help from the other views of that matrix, until all of the views are completed.

Our study is more similar to the works of \cite{Tsuda,Lanckriet,LanckrietC,Kato}, where different descriptions of yeast proteins are utilized (such as primary protein sequences, protein-protein interaction data, and mRNA expression data) for protein classification later on. Each description constitutes a data set. In contrast with their work, our study assumes that some or all of the descriptions (or data set) have missing rows and columns, where the set of proteins with missing information in each data set may not be the same. This is illustrated by the different positions of white rows and columns of the kernel matrices in Fig.~\ref{fig:setting}~(c). We can also see from Fig.~\ref{fig:setting} how our study~(c) differs from those of Tsuda \textit{et al.}~(a) and Kato \textit{et al.}~(b). The problem is now translated to multiple kernel matrix completion problem, which our study aims to solve.

In this paper, we present a solution to multiple kernel matrix completion problem by introducing an algorithm, which we call Mutual Kernel Matrix Completion (MKMC) algorithm, that exploits the EM algorithm to minimize the Kullback-Leibler (KL) divergences among the kernel matrices. The E-step and M-step are given in closed form \tslong{(the derivations of which are given in Appendix~\ref{sec:EM})},
making our method time- and memory-efficient.

The rest of the paper is organized as follows: The next section describes the setting for our proposed method; Sect.~\ref{sec:MKMC} describes the theory behind the proposed algorithm and how to implement it;
Sect.~\ref{sec:mkmcem} reviews the EM framework to show that the MKMC algorithm is an EM algorithm; 
Sect.~\ref{sec:Experiments} expounds the data used and experimental design; Sect.~\ref{sec:Results} discusses the empirical results; and Sect.~\ref{sec:Conclusion} concludes.
\tsshort{Some of proofs are refered to the longer version~\cite{Rivero-arxiv17}. }

\section{Problem Setting}
\label{sec:prob}
\begin{figure*}[t]
\begin{center}
  \begin{tabular}[t]{ll}
    \begin{tabular}{l}
    (a) Tsuda \textit{et al.} (2003) 
      \\
      \\
    \includegraphics[clip,scale=0.5]{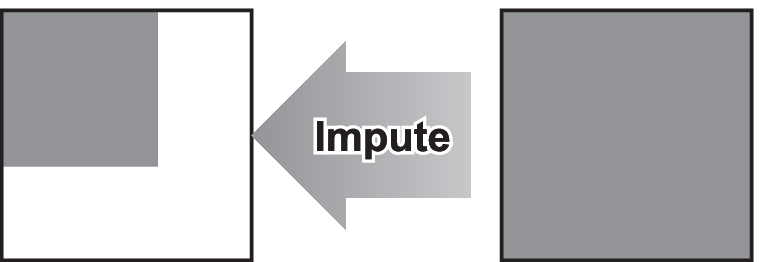}
    \\
    \vspace{0.5cm}
    \\
    (b) Kato \textit{et al.} (2005) 
    \\
    \\
    \includegraphics[clip,scale=0.5]{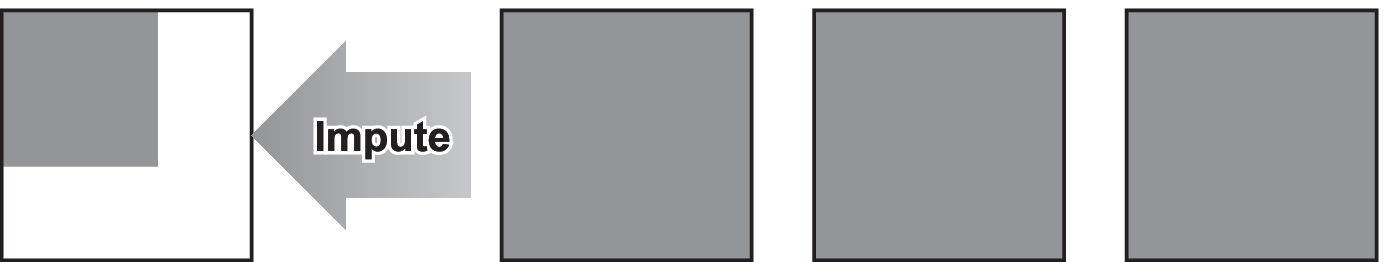}
    \end{tabular}
    &
    \begin{tabular}{l}
      (c) Ours
      \\
      \includegraphics[clip,scale=0.5]{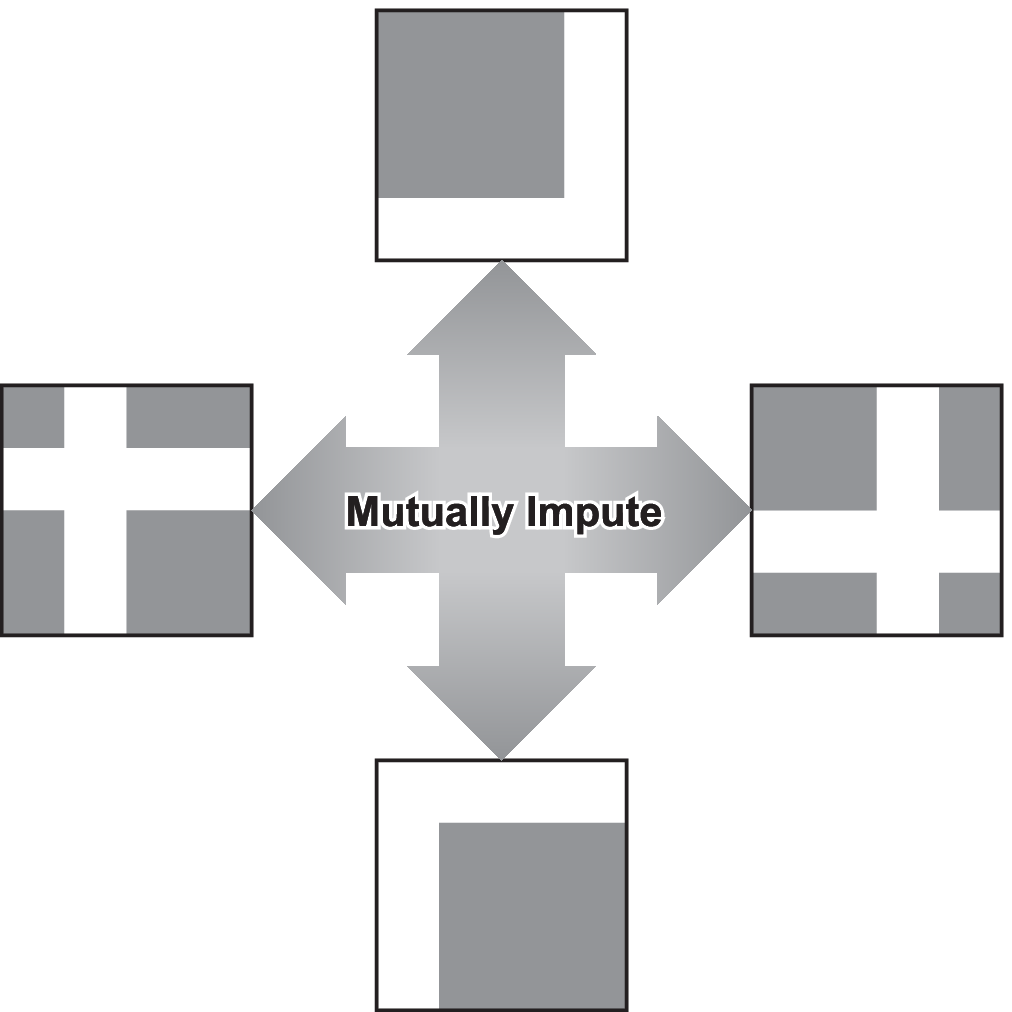}
    \end{tabular}
  \end{tabular}
\end{center}
\caption{Problem settings. 
Tsuda \textit{et al.}~\cite{Tsuda} used a single auxiliary complete matrix to complete a single incomplete matrix. Kato \textit{et al.}~\cite{Kato} used multiple auxiliary complete matrices to complete a single incomplete matrix. In this study, we mutually complete multiple incomplete matrices. }\label{fig:setting}
\end{figure*}

Suppose we wish to analyze $\ell$ objects and we have $K$ available relevant data sources for the same set of objects. The relationships among these objects in each of the data sources are given as entries in the following  $\ell\times \ell$ symmetric kernel matrices: $\left\{\boldsymbol{Q}^{(k)}\right\}_{k=1}^{K}$. Suppose also that some or all of the $K$ available data sources have missing entries, as can be seen in Fig.~\ref{fig:setting}~(c). The rows and columns with missing entries in each of the kernel matrices $\boldsymbol{Q}^{(k)}$ are rearranged such that the information is available only for the first $n_{k}<\ell$ samples in $\boldsymbol{Q}^{(k)}$, and unavailable for the remaining $m_{k} \defeq \ell\!-\!n_{k}$ samples. Then, the resultant reordered $\ell\times\ell$ matrix, denoted by $\boldsymbol{Q}^{(k)}_{vh,vh}$, can be partitioned as
\begin{equation}
	\arraycolsep=3pt
	\boldsymbol{Q}^{(k)}_{vh,vh} = \begin{pmatrix}
		\boldsymbol{Q}^{(k)}_{v,v}	&	\boldsymbol{Q}^{(k)}_{v,h} \cr
		\boldsymbol{Q}^{(k)}_{h,v}	&	\boldsymbol{Q}^{(k)}_{h,h} \cr
		\end{pmatrix},\label{eq:Q}
\end{equation}
where the $n_{k}\times n_{k}$ symmetric submatrix $\boldsymbol{Q}^{(k)}_{v,v}$ have visible entries, while the $n_{k}\times m_{k}$ submatrix $\boldsymbol{Q}^{(k)}_{v,h}$ and $m_{k}\times m_{k}$ submatrix $\boldsymbol{Q}^{(k)}_{h,h}$ have all entries missing. 

The goal of this study is to utilize the given information $\boldsymbol{Q}^{(1)}_{v,v}, \ldots, \boldsymbol{Q}^{(K)}_{v,v}$ to infer the missing information $\boldsymbol{Q}^{(1)}_{h,h}, \ldots, \boldsymbol{Q}^{(K)}_{h,h}$ and $\boldsymbol{Q}^{(1)}_{v,h}, \ldots, \boldsymbol{Q}^{(K)}_{v,h}$, thus completing the kernel matrices.

\section{Mutual Kernel Matrix Completion\tsshort{\\} (MKMC) Method}
\label{sec:MKMC}
To determine the difference between two kernel matrices, we will make use of the KL divergence between the corresponding probability distributions of the kernel matrices. To define the KL divergence, let us relate each of the kernel matrices to the covariance of a zero-mean Gaussian distribution as follows:
\begin{equation*}
	q_{1}\defeq \mathcal{N}\left(\textbf{0},\boldsymbol{Q}^{(1)} \right),\ldots, q_{K}\defeq \mathcal{N}\left( \textbf{0},\boldsymbol{Q}^{(K)}\right),
\end{equation*}
where the $q_{i}$'s are called the empirical distributions. Also, let us consider a model matrix $\boldsymbol{M}$ and relate it to the covariance of a zero-mean Gaussian distribution as $p \defeq \mathcal{N}\left(\textbf{0}, \boldsymbol{M}\right),$ called the model distribution. Then for $k=1,\ldots,K$, the KL divergence of probability distribution $p$ from $q_{k}$ is given by
\begin{equation*}
	KL\left(q_{k}, p\right)=\int\!q_{k}(\boldsymbol{x}) \log \dfrac{q_{k}(\boldsymbol{x})}{p(\boldsymbol{x})}\,d\boldsymbol{x},
\end{equation*}
and the KL divergence of the model matrix $\boldsymbol{M}$ from $\boldsymbol{Q}^{(k)}$ is given by
% \mathindent=3mm
\begin{eqnarray*}
\lefteqn{
	KL\left(\boldsymbol{Q}^{(k)}, \boldsymbol{M}\right)
}\quad \nonumber \\
	&=& \dfrac{1}{2}\left[\mbox{Tr}\left(\boldsymbol{M}^{-1}\boldsymbol{Q}^{(k)}\right)\!+\!\log\det\boldsymbol{M}\!-\!\log\det\boldsymbol{Q}^{(k)}\!-\!\ell\right],
\end{eqnarray*}
% \mathindent=7mm
where the size of $\boldsymbol{M}$ and $\boldsymbol{Q}^{(k)}$ is $\ell \!\times \!\ell$.
\tsshort{\newpage}
To infer the missing entries, we minimize the following objective function that takes the sum of the KL divergences:
% \mathindent=3mm
\begin{equation}
J(\mathcal{H}, \boldsymbol{M}) \coloneqq \lambda KL\left(\boldsymbol{I}, \boldsymbol{M} \right)\! +\! \sum_{k = 1}^{K} KL\left( \boldsymbol{Q}^{(k)}, \boldsymbol{M}\right),\label{eq:objftn}
\end{equation}
% \mathindent=7mm
where $\mathcal{H}\defeq \left\{\boldsymbol{Q}_{v,h}^{(k)},\boldsymbol{Q}_{h,h}^{(k)}\right\}_{k=1}^{K}$ is the set of submatrices with missing entries, and $\boldsymbol{M}$ is the model matrix.

Minimization of the KL divergence can be done by first fixing $\boldsymbol{M}$ and minimizing the objective function $J(\mathcal{H},\boldsymbol{M})$ with respect to $\boldsymbol{Q}^{(k)}_{v,h}$ and $\boldsymbol{Q}^{(k)}_{h,h}$, then fixing the obtained solutions to update $\boldsymbol{M}$.
Iteratively doing these steps
monotonically
% gradually
decreases the KL divergence of $\boldsymbol{M}$ from $\boldsymbol{Q}^{(k)}$, for each $k$.
These steps correspond to E-step and M-step in an EM algorithm, as detailed in the next section. %% These steps comprise the EM algorithm.

Succinctly, the following two steps are repeated until convergence:
\begin{enumerate}
\item \textit{E-step}. Fix the model matrix $\boldsymbol{M}$ to obtain $\mathcal{H}$.
\item \textit{M-step}. Use $\mathcal{H}$ to update $\boldsymbol{M}$.
\end{enumerate}
This algorithm is guaranteed to decrease the objective function monotonically. An outline of the proposed algorithm is given in Algorithm~\ref{alg:KMC}. In Step~7 of the algorithm, the model matrix $\boldsymbol{M}$ is reordered and partitioned in the same way as the current $\boldsymbol{Q}^{(k)}$. Since we are taking KL divergences from $\boldsymbol{M}$ to each of $\boldsymbol{Q}^{(k)}$s, the reordering of $\boldsymbol{M}$ as well as the size of its submatrices will depend on that of $\boldsymbol{Q}^{(k)}$ for which it is being compared to at the moment. We denote the reordered and partitioned $\boldsymbol{M}$ with respect to $\boldsymbol{Q}^{(k)}$ as
\begin{equation}
	\arraycolsep=3pt
	\boldsymbol{M}^{(k)}_{vh,vh} \coloneqq \begin{pmatrix}
		\boldsymbol{M}^{(k)}_{v,v}	&	\boldsymbol{M}^{(k)}_{v,h} \cr
		\boldsymbol{M}^{(k)}_{h,v}	&	\boldsymbol{M}^{(k)}_{h,h} \cr
		\end{pmatrix}.\label{eq:Mvhvh}
\end{equation}
We then denote the sizes for the submatrices as $n_{k}\times n_{k}$ for $\boldsymbol{M}^{(k)}_{v,v}$ and $\boldsymbol{Q}^{(k)}_{v,v}$; $m_{k}\times m_{k}$ for $\boldsymbol{M}^{(k)}_{h,h}$ and $\boldsymbol{Q}^{(k)}_{h,h}$; and $n_{k}\times m_{k}$ for $\boldsymbol{M}^{(k)}_{v,h}$ and $\boldsymbol{Q}^{(k)}_{v,h}$, where $m_{k}\defeq \ell\!-\!n_{k}$. $\boldsymbol{M}^{(k)}_{h,v}$ and $\boldsymbol{Q}^{(k)}_{h,v}$ denote the matrix transpose of the submatrices $\boldsymbol{M}^{(k)}_{v,h}$ and $\boldsymbol{Q}^{(k)}_{v,h}$, respectively.
\begin{algorithm}[th!]
\caption{
MKMC Algorithm.
\label{alg:KMC}}
\begin{algorithmic}[1]
  \REQUIRE
  Kernel matrices $\left\{\boldsymbol{Q}^{(k)}\right\}_{k = 1}^{K}$, some small positive number $\lambda$.
  \ENSURE Completed kernel matrices $\left\{\boldsymbol{Q}^{(k)}\right\}_{k = 1}^{K}$.
  \STATE \textbf{begin}
  \STATE Initialize $\left\{\boldsymbol{Q}^{(k)}\right\}_{k = 1}^{K}$ by imputing zeros in the missing entries;
  \STATE Initialize the model matrix as $\boldsymbol{M}=\frac{1}{\lambda\!+\!K} \left[\left(\sum_{k=1}^{K} \boldsymbol{Q}^{(k)}\right)\!+\!\lambda \boldsymbol{I}\right]$;\\
  \REPEAT
  \FORALL{$k \in \{1,\dots,K\}$}
  \STATE Reorder $\boldsymbol{Q}^{(k)}$ as $\boldsymbol{Q}_{vh,vh}^{(k)}$ and partition it as $\boldsymbol{Q}_{v,v}^{(k)}$, $\boldsymbol{Q}_{v,h}^{(k)}$, $\boldsymbol{Q}_{h,v}^{(k)}$, and $\boldsymbol{Q}_{h,h}^{(k)}$;
	\STATE Reorder $\boldsymbol{M}$ as $\boldsymbol{Q}_{vh,vh}^{(k)}$ and partition it as $\boldsymbol{M}_{v,v}^{(k)}$, $\boldsymbol{M}_{v,h}^{(k)}$, $\boldsymbol{M}_{h,v}^{(k)}$, and $\boldsymbol{M}_{h,h}^{(k)}$;
  \STATE $\boldsymbol{Q}_{v,h}^{(k)}\!\coloneqq\!\boldsymbol{Q}_{v,v}^{(k)}\left(\boldsymbol{M}_{v,v}^{(k)}\right)^{-1}\boldsymbol{M}_{v,h}^{(k)}$;
  \STATE $\boldsymbol{Q}_{h,h}^{(k)}\!\coloneqq\!\boldsymbol{M}_{h,h}^{(k)}\!-\!\boldsymbol{M}_{h,v}^{(k)}\left(\boldsymbol{M}_{v,v}^{(k)}\right)^{-1}\boldsymbol{M}_{v,h}^{(k)}+\boldsymbol{M}_{h,v}^{(k)}\left(\boldsymbol{M}_{v,v}^{(k)}\right)^{-1}\boldsymbol{Q}_{v,v}^{(k)}\left(\boldsymbol{M}_{v,v}^{(k)}\right)^{-1}\boldsymbol{M}_{v,h}^{(k)}$; 
  \ENDFOR
  \STATE Update $\boldsymbol{M}$ as $\boldsymbol{M}=\frac{1}{\lambda\!+\!K} \left[\left(\sum_{k=1}^{K} \boldsymbol{Q}^{(k)}\right)\!+\!\lambda \boldsymbol{I}\right].$\\  \UNTIL{convergence}
  \STATE \textbf{end.}
\end{algorithmic}
\end{algorithm}

\section{Connection to EM Algorithm}
\label{sec:mkmcem}
In this section, we review the EM framework and then show that MKMC is an EM algorithm.

\subsection{EM Algorithm Revisited}
The EM method is a numerical optimization algorithm for statistical inference of model parameters $\vThet$ such as maximum likelihood estimation. From observed data $\vV$, the maximum likelihood estimation finds the value of $\vThet$ that maximizes the log-likelihood function defined as 
\begin{align}
  L_{\text{ML}}(\vThet;\vV):=\log p(\vV\,|\,\vThet),
\end{align}
where $p(\vV\,|\,\vThet)$ is the model. 
A penalizing term is often added to the
log-likelihood function~\cite{Murphy12a}. 
The penalized log-likelihood function
is given as 
\begin{align}
  L_{\text{II}}(\vThet;\vV):= L_{\text{ML}}(\vThet;\vV)+ \log p(\vThet),
\end{align}
where $p(\vThet)$ is a predefined prior distribution
of the model parameters.
With an empirical distribution $q(\vV)$~\cite{Amari95a}, 
a slightly generalized form can be considered: 
\begin{align}
  L_{\text{II}}(\vThet;q)
  :=
  \bE_{q(\vV)}\left[\log p(\vV\,|\,\vThet)\right] + \log p(\vThet). 
\end{align}
Given the empirical distribution $q(\vV)$,
the EM framework iteratively
maximizes $L_{\text{II}}(\vThet;q)$ with respect to
$\vThet$ by repeating E-step and M-step alternately.
In the EM framework,
unobserved data $\vH$ are considered 
in addition to the observed data $\vV$,
and it is supposed that
the joint distribution, $p(\vV,\vH\,|\,\vThet)$,
is in the exponential
family~\cite{Wainwright08book}.
Namely, there exist vector-valued or matrix-valued functions,
$\vS(\cdot)$ and $\vG(\cdot)$, such that
\begin{align}
  \log p(\vV,\vH\,|\,\vThet)
  =
  \left<\vS(\vV,\vH),\vG(\vThet)\right> - A(\vThet),
\end{align}
where $A(\cdot)$ is known as the \emph{cumulant function}
\begin{align}
  A(\vThet)
  := \log \int \exp \left<\vS(\vV,\vH),\vG(\vThet)\right>d\vV d\vH.
\end{align}
The pair of $\vV$ and $\vH$ is called 
the \emph{complete data}~\cite{McLachlan08book}. 
The model distribution of the observed data $\vV$
is the marginal distribution of the joint distribution:  
\begin{align}
  p(\vV\,|\,\vThet)
  =
  \int p(\vV,\vH\,|\,\vThet) d\vH. 
\end{align}
In the E-step of $t$-th iterate,
the expected sufficient statistics
\begin{align}
  \vS^{(t)} := \bE_{t-1}
    \left[
      \vS(\vV,\vH)
    \right]
\end{align}
is computed where $\vThet^{(t-1)}$ is the value
determined in the previous iterate, and we denote
$\bE_{t-1} = \bE_{p(\vH\,|\,\vV,\vThet^{(t-1)})}\bE_{q(\vV)}$. 
Then,
in the M-step, the so-called Q-function~\cite{McLachlan08book}
is maximized with respect to $\vThet$ to get the new value
$\vThet^{(t)}$, where the Q-function is defined as
\begin{align}\label{eq:Q-def-general}
  \begin{split}
    Q^{(t)}(\vThet)
    &:=
    \bE_{t-1}\left[
      \log p(\vV,\vH\,|\,\vThet) 
      \right]
      + \log p(\vThet)
  \\
  &=
  \left<\vS^{(t)}, \vG(\vThet)\right> - A(\vThet)
  + \log p(\vThet).
  \end{split}
\end{align}
If the penalized log-likelihood function is bounded
from above, the convergence of the EM algorithm is
ensured from the following property: 
\begin{proposition}\label{prop:em-non-decreasing}
 The EM algorithm produces a non-decreasing series of the penalized log-likelihood:
$$L_{\text{II}}(\vThet^{(t)};q) \ge L_{\text{II}}(\vThet^{(t-1)};q),\text{ for $t\in\bN.$}$$ 
\end{proposition}
\noindent\tslong{(Proof is given in Appendix~\ref{ss:proof-em-non-decreasing}). }The proposed algorithm, MKMC, makes the solution
converge to a local optimum, and so does the EM algorithm. 

\subsection{MKMC is an EM Algorithm}
We are now ready to establish a connection between MKMC and the EM framework. In a setting described below, we can show that MKMC is an EM algorithm. We first describe the setting of the empirical distribution, followed by the definition of the model distribution. Now let us suppose the setting in which the complete data is expressed as a set of $\ell$-dimensional vectors $\x_{1},\dots,\x_{K}$ where each entry in $\x_{k}$ corresponds to a row or a column of $\vQ^{(k)}$.  For $k=1,\dots,K$, let $\vv_{k}\in\bR^{n_{k}}$ and $\vh_{k}\in\bR^{m_{k}}$ be the sub-vectors of $\x_{k}$, where the entries in $\vv_{k}$ correspond to available data in $k$-th kernel matrix, and the entries in $\vh_{k}$ correspond to missing data in $k$-th kernel matrix. In the context of EM framework, the observed data is $\vV:=(\vv_{1},\dots,\vv_{K})$ and the unobserved data is $\vH:=(\vh_{1},\dots,\vh_{K})$. The second order moment of the empirical distribution $q(\vV)$ is supposed to be given as
\begin{align}\label{eq:EvvT-is-Qvv}
  \bE_{q(\vV)}\left[\vv_{k}\vv_{k}^\top\right] = \vQ^{(k)}_{v,v}. 
\end{align}
Notice that an example that satisfies the assumption in \eqref{eq:EvvT-is-Qvv} is 
\begin{align}\label{eq:emp-is-prod-qk}
  q(\vV)=\int
  q_{1}(\vv_{1},\vh_{1})\cdot\cdots\cdot
  q_{K}(\vv_{K},\vh_{K}) d\vH, 
\end{align}
\noindent
\tslong{(proof is given in Appendix~\ref{ss:proof-emp-is-prod-qk}) }where $q_{1}$,\dots,$q_{K}$ have been defined in Sect.~\ref{sec:MKMC}, although the definition of the empirical distribution is not limited to \eqref{eq:emp-is-prod-qk} so long as the assumption in \eqref{eq:EvvT-is-Qvv} holds. 

Next, we define the model distribution. In the model, $\vThet = \vM$ and the joint densities of the complete data are given by
\begin{align}
  p(\vV,\vH\,|\,\vThet)
  =
  \prod_{k=1}^{K} \cN( \x_{k}\,|\,\0,\vM). 
\end{align}
Note that this model distribution is
in the exponential family by setting
\begin{align}\label{eq:ss-and-natupara-for-mkmcmdl}
  \vS(\vV,\vH) :=
  \sum_{k=1}^{K} \x_{k}\x_{k}^\top
  \quad\text{ and }\quad
  \vG(\vM) = -\frac{1}{2}\vM^{-1}\tslong{,}\tsshort{.} 
\end{align}
\tslong{the derivations of which are given in Appendix~\ref{ss:deriv-S-G}. }The prior distribution of $\vM$ is defined
as the following Wishart distribution:
\begin{align}
  p(\vM) \propto
  \left(
  \frac{1}{\text{det}\vM}
  \right)^{\lambda/2}
  \exp
  \left(
  -\frac{\lambda}{2}
  \text{Tr}
  \vM^{-1}
  \right). 
\end{align}
%
%\tsnote{Rachelle needs to replace $\epsilon$
% with $\lambda$.  Reviewer may thinks that
%  $\epsilon$ in the first submission
%  is a regularization parameter. }

We now apply the EM algorithm to this setting.
In the E-step, the expectation of the
sufficient statistics
\begin{align}
  \bE_{t-1}\left[\vS(\vV,\vH)\right] =
  \sum_{k=1}^{K}\bE_{t-1}\left[\x_{k}\x_{k}^\top\right]
\end{align}
has to be computed.
The $\ell\times\ell$ symmetric matrices
$\bE_{t-1}\left[\x_{k}\x_{k}^\top\right]$ have the following
sub-matrices:  
$\bE_{t-1}\left[\vv_{k}\vv_{k}^\top\right]$,
$\bE_{t-1}\left[\vv_{k}\vh_{k}^\top\right]$,
 and
$\bE_{t-1}\left[\vh_{k}\vh_{k}^\top\right]$. 
The $n_{k}\times n_{k}$ matrix
$\bE_{t-1}\left[\vv_{k}\vv_{k}^\top\right]$
does not depend on the model parameters~$\vM$. 
Letting
\begin{align}
 \vM^{(k)}_{h|v}
 :=
    \vM^{(k)}_{h,h}
    -\vM^{(k)}_{h,v}(\vM^{(k)}_{v,v})^{-1}
    \vM^{(k)}_{v,h}, 
\end{align}
it can be shown that the remaining two expected matrices are given by
\begin{eqnarray}
\arraycolsep=1pt
  \label{eq:E-step-of-vh-in-EM}
  \bE_{t-1}\left[\vv_{k}\vh_{k}^\top\right]\!&=&\!\vQ^{(k)}_{v,v}(\vM_{v,v}^{(k)})^{-1}\vM^{(k)}_{v,h}, \text{ and}
\end{eqnarray}
\tsshort{\newpage}
\begin{eqnarray}
\arraycolsep=1pt
  \label{eq:E-step-of-hh-in-EM}
	\lefteqn{
		\bE_{t-1}\left[\vh_{k}\vh_{k}^\top\right]=\vM^{(k)}_{h|v}
	}\quad \nonumber \\
  &+&\!\vM_{h,v}^{(k)}(\vM_{v,v}^{(k)})^{-1}
    \vQ^{(k)}_{v,v}(\vM^{(k)}_{v,v})^{-1}
    \vM^{(k)}_{v,h}, 
\end{eqnarray}
\noindent\tslong{(derivations are given in Appendix~\ref{ss:deriv-E-step-in-EM}) }which coincides with the update rule given in Sect.~\ref{sec:MKMC}.  Hence,
$\vQ^{(k)}$ obtained in the $t$-th iterate
in MKMC is equal to
$\bE_{t-1}\left[\x_{k}\x_{k}^\top\right]$. 

In the M-step, the Q-function is maximized
with respect to $\vM$. Substituting all the
settings to \eqref{eq:Q-def-general}, 
the Q-function is expressed as
\begin{align}
  \begin{split}
    Q^{(t)}(\vM)
  =&
  -\frac{1}{2}
  \left<\lambda\vI+\sum_{k=1}^{K}\vQ^{(k)},\vM^{-1}\right>
  \\
  &\qquad
  -\frac{K+\lambda}{2}\text{logdet}\vM
  + \text{const}.
  \end{split}
\end{align}
which is maximized at
\begin{align}
  \vM
  =
  \frac{1}{\lambda+K}
  \left(\lambda\vI+\sum_{k=1}^{K}\vQ^{(k)}\right).\label{eq:modelM}
\end{align}
It has turned out that
the update rules of the E-step and the M-step
are same as those of MKMC algorithm.
In conclusion, MKMC can be interpreted as
an EM algorithm.

\section{Experimental Results}
\label{sec:Experiments}

\subsection{Experimental Settings}
The MKMC algorithm is applied to seven kernel matrices derived from three different types of data: four from primary protein sequence, two from protein-protein interaction data, and one from mRNA expression data. These are the data used in \cite{Lanckriet} to test their statistical framework for genomic data fusion and yeast protein classification. They have shown in their experiment that the knowledge obtained from all of the aforementioned data improved protein classification, as compared to classifying proteins using only a single type of data. In order to utilize the heterogeneous data for protein classification task, the data are reduced into the common format of kernel matrices, allowing them to combine \cite{Lanckriet}. These kernel matrices are square symmetric matrices whose entries represent similarities among pairs of yeast proteins. Each gene is annotated as membrane~($+1$), non-membrane~($-1$), or unknown~($0$). In this study we only considered proteins with known annotations, leaving 2,318 yeast proteins out of 6,112.

In our experiment, for each object, one of seven kernel matrices is picked and the corresponding row and column from that matrix are ``removed'', i.e., replaced with undetermined values without changing the size of the matrix. We considered varying the ratio of the missing entries from 10\% up to 90\% missing entries. First, we randomly chose 10\% of the entries to be missed, where completion methods were to be performed. Next, we randomly chose another 10\% entries to be missed, in addition to the previous 10\% --- comprising the now 20\% missing entries. This was repeatedly done until we reached the 90\% missing entries. For comparison, other matrix completion techniques were considered such as zero-imputation and mean-imputation methods. Zero-imputation completes a kernel matrix by imputing zeros in its missing entries. This was also how we initialized the kernel matrices in the MKMC algorithm. On the other hand, imputing the mean of the remaining entries to complete a kernel matrix is called completion by mean-imputation. \tslong{Appendix}~\ref{sec:mean} describes how each of the kernel matrices was completed using this method.

The above-mentioned matrix completion techniques were individually performed on each of the seven kernel matrices. We then randomly picked data points to be used in training the SVM classifier for protein classification task afterwards.

In this study, 200 and 1,000 training data points were considered. We first made the division of training and testing datasets when 200 data are used for training, and then made the division when 1,000 training data are used. For the 200 training data, we randomly picked 200 out of 2,318 data points, and used them in training SVM classifiers in each of the kernel matrices. The remaining 2,118 data points served as testing data. The 1,000 training data points were obtained by retaining the first 200 training data points, and adding to it 800 more randomly-picked data points. This time, there were 1,318 testing data.

As regards to the completion methods, the MKMC algorithm gives us the completed kernel matrices $\left\{\boldsymbol{Q}^{(k)}\right\}_{k = 1}^{K}$, where $K$ is the number of kernel matrices used. SVM classifiers were then trained on these matrices, and on the combination of these kernel matrices, for the task of classifying a protein as membrane or non-membrane. Meanwhile, after completion by zero-imputation and mean-imputation, the kernel matrices were also combined to obtain the model matrix as in (\ref{eq:modelM}),
%\begin{equation*}
%	\boldsymbol{M}\!=\!\frac{1}{\lambda+K}\left[\sum_{k=1}^{K}\boldsymbol{Q}^{(k)}\!+\!\lambda\boldsymbol{I}\right], \label{eq:M}
%\end{equation*}
where the SVM classifier was also trained alongside with the seven completed kernel matrices. The zero-imputation and mean-imputation methods, applied with SVM, were then called zero-SVM and mean-SVM, respectively.
\begin{figure*}[t]
\begin{center}
  \begin{tabular}[t]{ll}
    \begin{tabular}{l}
    (a) Number of training examples: 200
     \\
    \includegraphics[clip,scale=0.55]{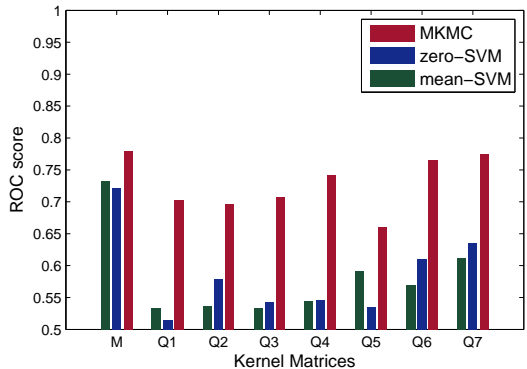}
    
    \end{tabular}
    &
    \begin{tabular}{l}
      (b) Number of training examples: 1,000
      \\
      \includegraphics[clip,scale=0.55]{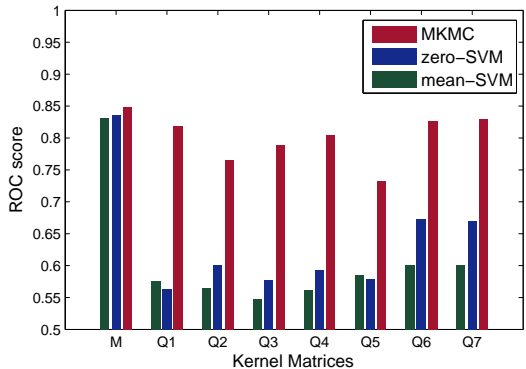}
    \end{tabular}
  \end{tabular}
\end{center}
\caption{ROC scores of the kernel matrix completion methods (MKMC, zero-SVM, and mean-SVM) on the model matrix M, and on each of the kernel matrices (Q1,...,Q7) when about $50\%$ of the entries were artificially missed. The number of training points considered were (a)~200 and (b)~1,000.}

%; the number of training points are 200 and 1,000; about $50\%$ of the entries were artificially missed, and were estimated by the three methods afterwards.}
\label{fig:ROC1}
\end{figure*}

\begin{figure}[t]
	\centering
		\includegraphics[width=0.47\textwidth]{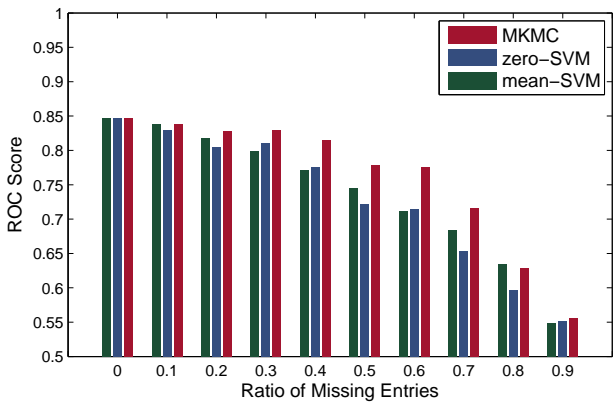}
	\caption{Classification performances of MKMC, zero-SVM, and mean-SVM on the model matrix $\boldsymbol{M}$ as the ratio of the missing values was artificially changed. Here, the number of training examples used was 200, and the parameter $\lambda$ was set to 0.001.}
	\label{fig:testforR_M_rocN200}
\end{figure}

\begin{figure}[t]
	\centering
		\includegraphics[width=0.47\textwidth]{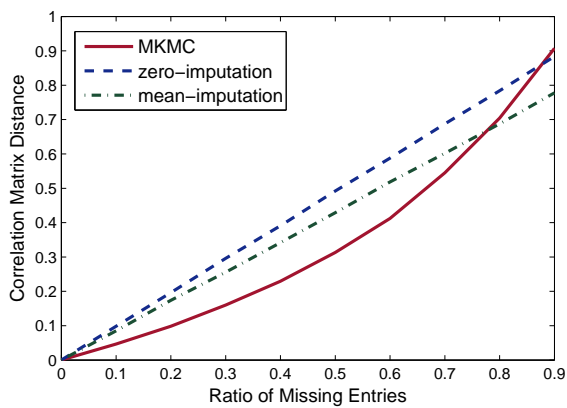}
	\caption{Test on how accurate the missing entries were recovered as the ratio of the missing values was artificially changed. The accuracy was measured by taking the mean of the correlation matrix distances between the estimated and their corresponding original kernel matrices.}
	\label{fig:MQd}
\end{figure}

%%\section{Results and Discussion}
\subsection{Results}
\label{sec:Results}
In this section, the classification performance and accuracy of the completion methods used were evaluated. The methods' classification performances were determined in different set-up: varying number of training points and varying ratio of missing entries. On the other hand, how accurate the methods were able to recover the missing data was evaluated by taking the mean of the correlation between the original matrices and the estimated matrices.

\subsubsection{Classification Performance}
The classification performance of the three completion methods were measured via the ROC scores. The ROC score, or area under the ROC curve, tells us how well a binary classifier can discriminate the data. Better classifiers achieve higher ROC scores. The ROC is also invariant of class label imbalance, which makes it the best tool in measuring the classification performance in our experiment since our data set is heavily imbalanced: the positively-labeled proteins are only 497 out of 2,318 (only $21.44\%$). In this experiment, we can determine how well a completion method preserves the relationship among pairs of proteins by measuring how accurate an SVM classifier can discriminate proteins as membrane or non-membrane. 

The ROC scores of the three completion methods on each kernel matrix, for 200 and 1,000 training points, are shown in Fig.~\ref{fig:ROC1}. Here, about $50\%$ of the entries in each of the kernel matrices were missing, and had been estimated by the three methods. Our empirical results show that as the number of training data increases, the SVM classification performance improves. We can also see from the plots that the model matrix $\boldsymbol{M}$, which is the combination of the completed kernel matrices, obtained the highest ROC score; this matches the claim by Lanckriet \textit{et al.} \cite{Lanckriet} that a learning algorithm performs better if trained from the integrated data than from a single data alone. Most importantly, our proposed method obtained the highest ROC scores in all of the classification performance tests, with significant differences compared to zero-SVM and mean-SVM. The statistical differences were assessed using a two-sample $t$-test. For 200 training data points, MKMC obtained an average ROC score of 0.779 for the model matrix $\boldsymbol{M}$, as compared to those of zero-SVM and mean-SVM: 0.721 (P-value $= 8.32 \cdot 10^{-5}$) and 0.745 (P-value $= 1.40 \cdot 10^{-3}$), respectively. On the other hand, when the number of training data is 1,000, MKMC achieved an average ROC score of 0.848 for $\boldsymbol{M}$ as compared to zero-SVM: 0.836 and mean-SVM: 0.832 (P-values $= 3.45 \cdot 10^{-2}$ and $4.60 \cdot 10^{-3}$, respectively).

In addition to the above tests, the effect of the amount of missing values in the kernel matrices on the methods' performances were also determined. The ratio of the missing entries was artificially changed from $0\%$ (no missing entries, as baseline) to $90\%$. Fig.~\ref{fig:testforR_M_rocN200} demonstrates the performances of MKMC, zero-SVM, and mean-SVM in this set-up. Here, although the classification performance decreased as the number of missing entries was increased, MKMC remained having the highest ROC scores among the three methods; the gap in the ROC scores being more prominent when there are $40\%$ to $70\%$ missing entries.

The experiments show that among the matrix completion techniques used, MKMC best preserves the relationships among the data points.

%----------------------------------------------------------------------------------------------------
\subsubsection{Completion Accuracy}
To test how accurate the missing values were recovered by the three methods, we computed the mean of the correlation distance between the $K$ estimated matrices and their corresponding complete kernel matrices as follows:
	\[ \dfrac{1}{K}\sum_{k=1}^{K}\left(1-\dfrac{\mbox{Tr}\left(\boldsymbol{Q}^{(k)}\hat{\boldsymbol{Q}}^{(k)}\right)}{\left\|\boldsymbol{Q}^{(k)}\right\|_{F}\left\|\hat{\boldsymbol{Q}}^{(k)}\right\|_{F}}\right),
\]
where $\left\|\cdot\right\|_{F}$ is the Frobenius norm, and $\boldsymbol{Q}^{(k)}$ and $\hat{\boldsymbol{Q}}^{(k)}$ are the ``true'' and the estimated kernel matrices, respectively; the computation was done for each ratio of missing entries. Lower correlation matrix distance means better correlation between the matrices. As with the previous experiments, the ratio of the missing entries was artificially changed from $0\%$ to $90\%$. Fig.~\ref{fig:MQd} shows how MKMC recovered the missing values more accurately than the two imputation methods, with the exception when about at least $80\%$ of the entries are missing. In this case, however, the gap between MKMC and mean-imputation is not very large.

%}
\section{Conclusion}
\label{sec:Conclusion}
In this study, we introduced a new algorithm, called MKMC, to solve the problem of mutually inferring the missing entries of similarity matrices (or kernel matrices). MKMC works by combining relevant data sources, although some or all are incomplete, and using the combined sources (called the model matrix) to estimate the missing entries of the kernel matrices. The inference is done by minimizing the KL divergence of the model matrix from each of the kernel matrices. The new kernel matrices are recombined, thus updating the model matrix which can be reused to improve the estimates of the entries. The optimal solutions given by MKMC in the E- and M-steps are given in closed forms, making MKMC efficient in terms of time and memory. In addition, results show that MKMC has good to excellent classification performance as the amount of training data and missing entries were varied--- a good indication that MKMC preserves relationships among pairs of data points. We have also shown that when portions of the kernel matrices were artificially missed, MKMC recovers the missing entries more accurately than the other two methods.
\section*{Acknowledgment}
This work was supported by JSPS KAKENHI Grant Number
26249075, 40401236.

\bibliographystyle{plain}
\bibliography{ieice2017rac-bib}

\begin{footnotesize}
\appendix
\tslong{
\section{Derivation of the Closed-Form Solutions}
\label{sec:EM}
This section shows us how the closed-form solutions to the MKMC algorithm were obtained.

To begin with, let us introduce an objective function $J$ that takes the sum of the KL divergences:
% \methindent=0mm
\begin{eqnarray}
J(\mathcal{H}, \boldsymbol{M}) &\coloneqq& \lambda KL\left(\boldsymbol{I}, \boldsymbol{M} \right)\! +\! \sum_{k = 1}^{K} KL\left( \boldsymbol{Q}^{(k)}, \boldsymbol{M}\right)\nonumber\\
&=& \dfrac{\lambda}{2}\mbox{Tr}\left(\boldsymbol{M}^{-1}\right)\!+\!\dfrac{\lambda}{2}\log\det\boldsymbol{M}\!-\!\dfrac{\lambda}{2}\ell \nonumber\\
& & \mbox{}+\!\dfrac{1}{2}\sum_{k=1}^{K}\mbox{Tr}\left(\boldsymbol{M}^{-1}\boldsymbol{Q}^{(k)}\right)\!+\!\dfrac{K}{2}\log\det\boldsymbol{M} \nonumber\\
& & \mbox{}-\!\dfrac{1}{2}\sum_{k=1}^{K}\log\det\boldsymbol{Q}^{(k)}\!-\!\dfrac{K}{2}\ell\nonumber\\
&=& \dfrac{\lambda}{2}\mbox{Tr}\left(\boldsymbol{M}^{-1}\right)\!+\!\dfrac{\lambda\!+\!K}{2}\log\det\boldsymbol{M}\nonumber\\
& & \mbox{}+\!\dfrac{1}{2}\sum_{k=1}^{K}\left[\mbox{Tr}\left(\boldsymbol{M}^{-1}\boldsymbol{Q}^{(k)}\right)\!-\!\log\det\boldsymbol{Q}^{(k)}\right]\nonumber\\
& & \mbox{}-\!\left(\lambda\!+\!K\right)\dfrac{\ell}{2},\label{eq:JJ}
\end{eqnarray}
% \methindent=7mm
where $\boldsymbol{Q}^{(1)},\boldsymbol{Q}^{(2)},\ldots,\boldsymbol{Q}^{(K)}$ are the $K$ data sources represented as $\ell\times\ell$ symmetric kernel matrices in which some or all may be incomplete; $\boldsymbol{M}$ is the model matrix; and $\lambda$ is a small positive scalar.

%%%%%%%%%%%%%%%%%%%%%%%%%%%%%%%%%%%%%%%%%%%%%%%%%%%%%%%%%%%%
\subsection{E-step.} In this step, we fix $\boldsymbol{M}$ and minimize $J$ with respect to $\mathcal{H}$.

First, suppose $\boldsymbol{Q}^{(k)}$ and $\boldsymbol{M}$ are reordered and partitioned as in (\ref{eq:Q})~and~(\ref{eq:Mvhvh}), respectively. By Schur \cite{Schur}, the Schur complement of $\boldsymbol{M}^{(k)}_{vh,vh}$ with respect to $\boldsymbol{M}^{(k)}_{v,v}$ is defined as
\begin{equation*}
	\boldsymbol{M}^{(k)}_{vh,vh}/\boldsymbol{M}^{(k)}_{v,v}\defeq\boldsymbol{M}^{(k)}_{h,h}\!-\!\boldsymbol{M}^{(k)}_{h,v}\left(\boldsymbol{M}^{(k)}_{v,v}\right)^{-1}\boldsymbol{M}^{(k)}_{v,h},
\end{equation*}
and the determinant of $\boldsymbol{M}^{(k)}_{vh,vh}$ is given by
\begin{equation*}
	\det\boldsymbol{M}^{(k)}_{vh,vh}=\det\boldsymbol{M}^{(k)}_{v,v}\det\left(\boldsymbol{M}^{(k)}_{vh,vh}/\boldsymbol{M}^{(k)}_{v,v}\right).
\end{equation*}
Hence,
% \methindent=0mm
\begin{eqnarray}
\lefteqn{
	\log\det\boldsymbol{M}^{(k)}_{vh,vh}
	}\quad \nonumber \\
	&=&\log\det\boldsymbol{M}^{(k)}_{v,v}\nonumber\\
	& &\mbox{}+\!\log\det\left[\boldsymbol{M}^{(k)}_{h,h}\!-\!\boldsymbol{M}^{(k)}_{h,v}\left(\boldsymbol{M}^{(k)}_{v,v}\right)^{-1}\!\boldsymbol{M}^{(k)}_{v,h}\right].\label{eq:MM}
\end{eqnarray}
% \methindent=7mm
Similarly for each $k$ in $\boldsymbol{Q}^{(k)}_{vh,vh}$,
% \methindent=0mm
\begin{eqnarray}
\lefteqn{
	\log\det\boldsymbol{Q}_{vh,vh}^{(k)}
	}\quad \nonumber \\
	&=&\log\det\boldsymbol{Q}^{(k)}_{v,v}\nonumber\\
	& &\mbox{}+\!\log\det\left[\boldsymbol{Q}^{(k)}_{h,h}\!-\!\boldsymbol{Q}^{(k)}_{h,v}\left(\boldsymbol{Q}^{(k)}_{v,v}\right)^{-1}\boldsymbol{Q}^{(k)}_{v,h}\right].\label{eq:QQ}
\end{eqnarray}
% \methindent=7mm
Let us now represent the inverse of $\boldsymbol{M}^{(k)}_{vh,vh}$ as
\begin{equation*}
	\arraycolsep=3pt
	\left(\boldsymbol{M}_{vh,vh}^{(k)}\right)^{-1} = \begin{pmatrix}
		\boldsymbol{S}^{(k)}_{v,v}	&	\boldsymbol{S}^{(k)}_{v,h} \cr
		\boldsymbol{S}^{(k)}_{h,v}	&	\boldsymbol{S}^{(k)}_{h,h} \cr
	\end{pmatrix} \eqqcolon \boldsymbol{S}^{(k)},
\end{equation*}
where
\begin{description}
	\item $\boldsymbol{S}^{(k)}_{v,v} = \left(\boldsymbol{M}_{v,v}^{(k)}\right)^{-1}\\
	 \ \ \ \ \ \ \ \ \ \ \ \ \ +\left(\boldsymbol{M}_{v,v}^{(k)}\right)^{-1}\boldsymbol{M}^{(k)}_{v,h}\left(\boldsymbol{M}^{(k)}_{vh,vh}/\boldsymbol{M}^{(k)}_{v,v}\right)^{-1}\boldsymbol{M}^{(k)}_{h,v}\left(\boldsymbol{M}^{(k)}_{v,v}\right)^{-1}$;
	\item $\boldsymbol{S}^{(k)}_{v,h} = -\left(\boldsymbol{M}^{(k)}_{v,v}\right)^{-1}\boldsymbol{M}^{(k)}_{v,h}\left(\boldsymbol{M}^{(k)}_{vh,vh}/\boldsymbol{M}^{(k)}_{v,v}\right)^{-1}$;
	\item $\boldsymbol{S}^{(k)}_{h,v} = -\left(\boldsymbol{M}^{(k)}_{vh,vh}/\boldsymbol{M}^{(k)}_{v,v}\right)^{-1}\boldsymbol{M}^{(k)}_{h,v}\left(\boldsymbol{M}^{(k)}_{v,v}\right)^{-1}$;
	\item $\boldsymbol{S}^{(k)}_{h,h} = \left(\boldsymbol{M}^{(k)}_{vh,vh}/\boldsymbol{M}^{(k)}_{v,v}\right)^{-1}$.
\end{description}
Then
% \methindent=3mm
\begin{eqnarray}
	\mbox{Tr}\left[\left(\boldsymbol{M}^{(k)}_{vh,vh}\right)^{-1}\boldsymbol{Q}^{(k)}_{vh,vh}\right]&=&\mbox{Tr}\left(\boldsymbol{S}^{(k)}_{v,v}\boldsymbol{Q}^{(k)}_{v,v}\right)+2\mbox{Tr}\left(\boldsymbol{S}^{(k)}_{v,h}\boldsymbol{Q}^{(k)}_{v,h}\right)\nonumber\\
	& &\mbox{}+\!\mbox{Tr}\left(\boldsymbol{S}^{(k)}_{h,h}\boldsymbol{Q}^{(k)}_{h,h}\right).\label{eq:MQ}
\end{eqnarray}
% \methindent=7mm
Rewriting the objective function $J$ in terms of the reordered matrices and of (\ref{eq:MM}),~(\ref{eq:QQ}), and (\ref{eq:MQ}), and taking its partial derivative with respect to $\boldsymbol{Q}^{(k)}_{h,h}$, we get
% \methindent=0mm
\begin{eqnarray}
	\dfrac{\partial J}{\partial \boldsymbol{Q}^{(k)}_{h,h}}&=&\dfrac{\partial\left[\displaystyle\sum_{k=1}^{K}\mbox{Tr}\left(\boldsymbol{S}^{(k)}_{h,h}\boldsymbol{Q}^{(k)}_{h,h}\right)\right]}{2\partial \boldsymbol{Q}^{(k)}_{h,h}}\nonumber\\	
	& &\mbox{}-\dfrac{\partial\left[\displaystyle\sum_{k=1}^{K}\log\det\left(\boldsymbol{Q}^{(k)}_{h,h}\!-\!\boldsymbol{Q}^{(k)}_{h,v}\left(\boldsymbol{Q}^{(k)}_{v,v}\right)^{-1}\boldsymbol{Q}^{(k)}_{v,h}\right)\right]}{2\partial \boldsymbol{Q}^{(k)}_{h,h}}\nonumber\\	
	&=&\dfrac{1}{2}\boldsymbol{S}^{(k)}_{h,h}\!-\!\dfrac{1}{2}\left[\left( \boldsymbol{Q}^{(k)}_{h,h}\!-\!\boldsymbol{Q}^{(k)}_{h,v}\left(\boldsymbol{Q}^{(k)}_{v,v}\right)^{-1}\boldsymbol{Q}^{(k)}_{v,h}\right)^{-1} \right]^{\top},\nonumber
\end{eqnarray}
% \methindent=7mm
for $k=1,\ldots,K$. Equating the above equation to \textbf{0} and solving for $\boldsymbol{Q}^{(k)}_{h,h}$ , we get
\begin{equation}
	\boldsymbol{Q}^{(k)}_{h,h}=\left(\boldsymbol{S}_{h,h}^{(k)}\right)^{-1}\!+\!\boldsymbol{Q}^{(k)}_{h,v}\left(\boldsymbol{Q}^{(k)}_{v,v}\right)^{-1}\boldsymbol{Q}^{(k)}_{v,h}.\label{eq:Qhh}
\end{equation}
Next, we substitute (\ref{eq:Qhh}) to (\ref{eq:MQ}) and solve for $\dfrac{\partial J}{\partial \boldsymbol{Q}^{(k)}_{v,h}}=\boldsymbol{0}$ for $k=1,\ldots,K$. As a result, we have
\begin{equation*}
	\boldsymbol{S}^{(k)}_{h,v}+\frac{1}{2}\boldsymbol{S}^{(k)}_{h,h}\left[2\boldsymbol{Q}^{(k)}_{h,v}\left(\boldsymbol{Q}^{(k)}_{v,v}\right)^{-1}\right]=\textbf{0}.
\end{equation*}
Solving for $\boldsymbol{Q}^{(k)}_{h,v}$ and taking its transpose, we obtain
\begin{equation}
	\boldsymbol{Q}^{(k)}_{v,h}=\mbox{}-\boldsymbol{Q}^{(k)}_{v,v}\boldsymbol{S}^{(k)}_{v,h}\left(\boldsymbol{S}_{h,h}^{(k)}\right)^{-1},\label{eq:Qvh}
\end{equation}
which we will substitute to (\ref{eq:Qhh}) to get
\begin{equation}
	\boldsymbol{Q}^{(k)}_{h,h}=\left(\boldsymbol{S}_{h,h}^{(k)}\right)^{-1}\!+\!\left(\boldsymbol{S}_{h,h}^{(k)}\right)^{-1}\boldsymbol{S}_{h,v}^{(k)}\boldsymbol{Q}^{(k)}_{v,v}\boldsymbol{S}_{v,h}^{(k)}\left(\boldsymbol{S}_{h,h}^{(k)}\right)^{-1}.\label{eq:Qhhh}
\end{equation}
Rewriting (\ref{eq:Qvh}) and (\ref{eq:Qhhh}) in terms of the submatrices of $\boldsymbol{Q}_{vh,vh}^{(k)}$ and $\boldsymbol{M}_{vh,vh}^{(k)}$, we arrive at
\begin{eqnarray*}
	\boldsymbol{Q}_{v,h}^{(k)}&=&\boldsymbol{Q}_{v,v}^{(k)}\left(\boldsymbol{M}_{v,v}^{(k)}\right)^{-1} \boldsymbol{M}^{(k)}_{v,h};\\
	\boldsymbol{Q}_{h,h}^{(k)}&=&\boldsymbol{M}^{(k)}_{h,h}\!-\!\boldsymbol{M}^{(k)}_{h,v}\left(\boldsymbol{M}_{v,v}^{(k)}\right)^{-1}\boldsymbol{M}^{(k)}_{v,h}\\
	& &\mbox{}+\!\boldsymbol{M}^{(k)}_{h,v}\left(\boldsymbol{M}_{v,v}^{(k)}\right)^{-1}\boldsymbol{Q}_{v,v}^{(k)}\left(\boldsymbol{M}^{(k)}_{v,v}\right)^{-1}\boldsymbol{M}^{(k)}_{v,h},
\end{eqnarray*}
which gives us the closed-form solutions for the E-step.

%%%%%%%%%%%%%%%%%%%%%%%%%%%%%%%%%%%%%%%%%%%%%%%%%%%%%%%%%%%%
\subsection{M-step.} In this step, $\mathcal{H}$ is fixed and we define
% \methindent=0mm
\begin{eqnarray*}
	J\left(\mathcal{H},\boldsymbol{S}^{-1}\right)&=&\dfrac{\lambda}{2}\mbox{Tr}\left(\boldsymbol{S}\right)+\dfrac{1}{2}\sum_{k=1}^{K}\left[\mbox{Tr}\left(\boldsymbol{S}\boldsymbol{Q}^{(k)}\right)-\log\det\boldsymbol{Q}^{(k)}\right]\\
	& &\mbox{}-\dfrac{\lambda+K}{2}\log\det\boldsymbol{S}-\dfrac{\lambda+K}{2}\ell \eqqcolon J'\left(\boldsymbol{S}\right).
\end{eqnarray*}
% \methindent=7mm
$J'$ is minimized when $\nabla_{\boldsymbol{S}}J'=\textbf{0}$, that is, when
\begin{equation*}
	\dfrac{\lambda}{2}\boldsymbol{I}+\dfrac{1}{2}\sum_{k=1}^{K}\boldsymbol{Q}^{(k)}-\dfrac{\lambda+K}{2}\boldsymbol{S}^{-1}=\textbf{0}.
\end{equation*}
Solving for $\boldsymbol{S}$, we get
% \methindent=0mm
\begin{equation}
	\boldsymbol{S}=\left(\dfrac{1}{\lambda\!+\!K}\left[\sum_{k=1}^{K}\boldsymbol{Q}^{(k)}\!+\!\lambda\boldsymbol{I}\right]\right)^{-1}=\begin{aligned}
				& \underset{\boldsymbol{S}\in\mathbb{S}_{++}^{\ell}}{\text{argmin}}\!\!\!\!\!
				& J'(\boldsymbol{S}),\end{aligned}\label{eq:minimizer}
\end{equation}
% \methindent=7mm
where $\mathbb{S}_{++}^{\ell}$ denotes the set of $\ell\times\ell$ positive-definite symmetric matrices. Since matrix inverse is one-to-one, we can express $\boldsymbol{S}^{-1}$ as $\boldsymbol{M}$, thus, from (\ref{eq:minimizer}),
\begin{equation*}
	\boldsymbol{M}=\dfrac{1}{\lambda\!+\!K}\left[\sum_{k=1}^{K}\boldsymbol{Q}^{(k)}\!+\!\lambda\boldsymbol{I}\right]
\end{equation*}
minimizes $J\left(\mathcal{H},\boldsymbol{S}^{-1}\right)=J\left(\mathcal{H},\boldsymbol{M}\right)$, giving us the closed-form solution for the M-step.

}
\section{The Mean-Imputation Method}
\label{sec:mean}
This section describes how each of the kernel matrices are completed using mean-imputation method.

Let us again partition the kernel matrices $\boldsymbol{Q}^{(k)}$ (for $k=1,\ldots,K$) as shown in (\ref{eq:Q}), with the corresponding sizes for the submatrices. Let us also denote the entries for $\boldsymbol{Q}^{(k)}_{v,v}$ as $K\left(\boldsymbol{x}_{i},\boldsymbol{x}_{j}\right)=\left\langle \boldsymbol{x}_{i},\boldsymbol{x}_{j}\right\rangle,$ where $\left\langle \cdot , \cdot\right\rangle$ denotes the inner product. The entries for the submatrix $\boldsymbol{Q}^{(k)}_{h,h}$ will then be given by 
\begin{equation*}
	K\left(\bar{\boldsymbol{x}},\bar{\boldsymbol{x}}\right)=\left\langle \bar{\boldsymbol{x}},\bar{\boldsymbol{x}}\right\rangle=\dfrac{1}{\left|I_{v}\right|^{2}}\sum_{i\in I_{v}}\sum_{j\in I_{v}}\left\langle \boldsymbol{x}_{i},\boldsymbol{x}_{j}\right\rangle,
\end{equation*}
where	$\bar{\boldsymbol{x}}=\dfrac{1}{\left|I_{v}\right|}\displaystyle\sum_{i\in I_{v}}\boldsymbol{x}_{i}$, and $I_{v}$ denotes the index set of the visible entries $\left(\text{i.e., the entries of }\boldsymbol{Q}^{(k)}_{v,v}\right)$. On the other hand, the entries for the submatrix $\boldsymbol{Q}^{(k)}_{v,h}$ is given by
\begin{equation*}
	K\left(\boldsymbol{x}_{j},\bar{\boldsymbol{x}}\right)=\left\langle \boldsymbol{x}_{j},\bar{\boldsymbol{x}}\right\rangle=\dfrac{1}{\left|I_{v}\right|}\sum_{i\in I_{v}}\left\langle \boldsymbol{x}_{j},\boldsymbol{x}_{i}\right\rangle,
\end{equation*}
while the submatrix $\boldsymbol{Q}^{(k)}_{h,v}$ is the matrix transpose of $\boldsymbol{Q}^{(k)}_{v,h}$. We have now obtained the values that complete the kernel matrices using mean-imputation method.

\tslong{
\section{Remaining Proofs and Derivations}
\subsection{Proof of Proposition~\ref{prop:em-non-decreasing}}
\label{ss:proof-em-non-decreasing}
We shall use the inequality:
\begin{align}
  \begin{split}\label{eq:gibs-ineq}
  &\forall \vThet, \  \bE_{t-1}\left[\log p(\vH\,|\,\vV,\vThet)\right]
  \\
  &\quad \le
  \bE_{t-1}\left[\log p(\vH\,|\,\vV,\vThet^{(t-1)})\right]
  \end{split}
\end{align}
which is immediately derived from the non-negativity
of the KL divergence: 
\begin{align}
  \text{KL}( p(\cdot\,|\,\vV,\vThet^{(t-1)}), p(\cdot\,|\,\vV,\vThet ) )
  \ge 0. 
\end{align}
Under the assumption that
$p(\vH\,|\,\vV,\vThet)>0$, 
the equality
\begin{align}
  \log p(\vV\,|\,\vThet)
  =
  \log p(\vV,\vH\,|\,\vThet)
  -\log p(\vH\,|\,\vV,\vThet) 
\end{align}
holds for any value of unobserved data $\vH$.
Taking the expectation according to 
$q(\vV)$ and $p(\vH\,|\,\vV,\vThet^{(t-1)})$ and adding
the penalizing term $\log p(\vThet)$,
the LHS becomes $L_{\text{II}}(\vThet;q)$. Hence, 
we get
\begin{align}
  L_{\text{II}}(\vThet;q)
  =
  Q^{(t)}(\vThet)
  -
  \bE_{t-1}\left[\log p(\vH\,|\,\vV,\vThet)\right]. 
\end{align}
From the update rule of the M-step,
we have 
$Q^{(t)}(\vThet^{(t)}) \ge Q^{(t)}(\vThet^{(t-1)})$.
With help of this and \eqref{eq:gibs-ineq}, we get
\begin{align}
  \begin{split}
  &L_{\text{II}}(\vThet^{(t)};q)
  =
  Q^{(t)}(\vThet^{(t)})
  -
  \bE_{t-1}\left[\log p(\vH\,|\,\vV,\vThet^{(t)})\right]
  \\
  &\ge 
  Q^{(t)}(\vThet^{(t-1)})
  -
  \bE_{t-1}\left[\log p(\vH\,|\,\vV,\vThet^{(t-1)})\right]
  \\
  &=
  L_{\text{II}}(\vThet^{(t-1)};q). 
  \end{split}
\end{align}

\subsection{\eqref{eq:EvvT-is-Qvv} holds with \eqref{eq:emp-is-prod-qk}}
\label{ss:proof-emp-is-prod-qk}
In the setting of \eqref{eq:emp-is-prod-qk},
note that 
\begin{align}
  q_{k}(\vv_{k}) = \cN(\vv_{k}\,|\,\0,\vQ^{(k)}_{v,v}) 
\end{align}
from the definition of $q_{k}$. 
The empirical distribution $q(\vV)$ is expressed
as the product of those $K$ Gaussians: 
\begin{align}
  \begin{split}
  q(\vV)
  &=
  \prod_{k=1}^{K}\int q_{k}(\vv_{k},\vh_{k})d\vh_{k}
  =
  \prod_{k=1}^{K}q_{k}(\vv_{k})
  \\
  &=
  \prod_{k=1}^{K}\cN(\vv_{k}\,|\,\0,\vQ^{(k)}_{v,v}). 
  \end{split}
\end{align}
From this, we have
\begin{align}
  \bE_{q(\vV,\vH)}\left[\vv_{k}\vv_{k}^\top\right] =
  \bE_{q_{k}(\vv_{k})}\left[\vv_{k}\vv_{k}^\top\right] =
  \vQ^{(k)}_{v,v}. 
\end{align}

\subsection{Derivation of \eqref{eq:ss-and-natupara-for-mkmcmdl}}
\label{ss:deriv-S-G}
\begin{align}
  \begin{split}
    &p(\vV,\vH\,|\,\vThet)=\prod_{k=1}^{K} \cN( \x_{k}\,|\,\0,\vM)
    \\
    &=
    \prod_{k=1}^{K}
    \frac{1}{(2\pi)^{\ell/2}\sqrt{\text{det}(\vM)}}
    \exp\left(
    -\frac{1}{2}\left<\x_{k},\vM^{-1}\x_{k}\right>
    \right)
    \\
    &=
    \frac{1}{(2\pi)^{\ell K/2}\left(\text{det}(\vM)\right)^{K/2}}    
    \exp\left(
    \left<\sum_{k=1}^{K}\x_{k}\x_{k}^\top,-\frac{1}{2}\vM^{-1}\right>
    \right)
    \\
    &=
    \frac{1}{(2\pi)^{\ell K/2}\left(\text{det}(\vM)\right)^{K/2}}    
    \exp\left(
    \left<\vS(\vV,\vH),\vG(\vM)\right>
    \right).
  \end{split}
\end{align}
\subsection{Derivations of \eqref{eq:E-step-of-vh-in-EM} and 
  \eqref{eq:E-step-of-hh-in-EM}}
\label{ss:deriv-E-step-in-EM}
Using
\begin{align}
  p(\vh_{k}\,|\,\vv_{k},\vM)
  =
  \cN\left( \vh_{k}\,|\,
  \vM^{(k)}_{h,v}(\vM^{(k)}_{v,v})^{-1}\vv_{k},\,\vM^{(k)}_{h|v}  
  \right),
\end{align}
\eqref{eq:E-step-of-vh-in-EM} is derived as
\begin{align}
  \begin{split}
    &\bE_{t-1}\left[\vv_{k}\vh_{k}^\top\right]
    =
  \bE_{q(\vV)}
  \left[
    \vv_{k}\bE_{p(\vH\,|\,\vV,\vThet^{(t-1)})}
    \left[\vh_{k}^\top\right]\right]
  \\
  &=\bE_{q(\vV)}
  \left[
    \vv_{k}
    \vv_{k}^\top(\vM^{(k)}_{v,v})^{-1}\vM^{(k)}_{v,h}
    \right]
  \\
  &=
\bE_{q(\vV)}
  \left[
    \vv_{k}
    \vv_{k}^\top
    \right]
  (\vM^{(k)}_{v,v})^{-1}\vM^{(k)}_{v,h}
  =
  \vQ^{(k)}_{v,v}(\vM^{(k)}_{v,v})^{-1}\vM^{(k)}_{v,h},
  \end{split}
\end{align}
and
\eqref{eq:E-step-of-hh-in-EM} is derived as
\begin{align}
  \begin{split}
    &\bE_{t-1}\left[\vh_{k}\vh_{k}^\top\right]
    \\
    &=
  \bE_{q(\vV)}
  \left[
    \vM^{(k)}_{h|v}
    +
    \vM^{(k)}_{h,v}(\vM^{(k)}_{v,v})^{-1}
    \vv_{k}\vv_{k}^\top
    (\vM^{(k)}_{v,v})^{-1}
    \vM^{(k)}_{v,h}\right]
  \\
  &=
    \vM^{(k)}_{h|v}
    +
    \vM^{(k)}_{h,v}(\vM^{(k)}_{v,v})^{-1}
    \vQ^{(k)}_{v,v}
     (\vM^{(k)}_{v,v})^{-1}
    \vM^{(k)}_{v,h}.
  \end{split}
\end{align}

}
%% \appendix*
\tsshort{
  \input{ieice2017rac-210-profile}
  }
\end{footnotesize}

\end{document}